\documentclass[12pt,onecolumn]{article}
\usepackage[latin1]{inputenc}
\usepackage{amsmath,hyperref,amsfonts,amssymb,graphicx,authblk,placeins,cite,color}
\usepackage{comment}

\begin{document}

\pagenumbering{gobble}

\title{Will we ever have Conscious Machines?}

\author[1,2]{Patrick Krauss}
\author[3]{Andreas Maier}
\affil[1]{\small Neuroscience Lab, Experimental Otolaryngology, University Hospital Erlangen, Germany}
\affil[2]{\small Cognitive Computational Neuroscience Group at the Chair of English Philology and Linguistics, Department of English and American Studies, Friedrich-Alexander University Erlangen-N\"urnberg (FAU), Germany}
\affil[3]{\small Chair of Pattern Recognition, Friedrich-Alexander University Erlangen-N\"urnberg (FAU), Germany}

\maketitle

\noindent\textbf{Corresponding author:} \\
Prof. Andreas Maier  \\
Chair of Pattern Recognition \\
Friedrich-Alexander University of Erlangen-N\"urnberg (FAU), Germany \\
E-Mail: andreas.maier@fau.de \\ \\

\noindent\textbf{Keywords:} \\
machine consciousness, artificial intelligence, theories of consciousness \\ \\ \\

\newpage
\begin{abstract}
The question of whether artificial beings or machines could become self-aware or consciousness has been a philosophical question for centuries. The main problem is that self-awareness cannot be observed from an outside perspective and the distinction of whether something is really self-aware or merely a clever program that pretends to do so cannot be answered without access to accurate knowledge about the mechanism's inner workings. We review the current state-of-the-art regarding these developments and investigate common machine learning approaches with respect to their potential ability to become self-aware. We realise that many important algorithmic steps towards machines with a core consciousness have already been devised. For human-level intelligence, however, many additional techniques have to be discovered.
\end{abstract}

\section*{Introduction}

The question of understanding consciousness is in the focus of philosophers and researchers for more than two millennia. Insights range broadly from \emph{``Ignorabimus'' -- ``We will never know.''}\footnote{With this simple statement, Emil du Bois-Reymond concluded his talk on the limits of scientific knowledge about the relation of brain processes and subjective experience at the 45th annual meeting of German naturalists and physicians in 1872.} to mechanistic ideas with the aim to construct artificial consciousness following 
Richard Feynman's famous words \emph{``What I cannot create, I do not understand.''}\footnote{Richard Feynman left these words on his blackboard in 1988 at the time of his death as a final message to the world.}.

The major issue that precludes the analysis of consciousness is its subjectivity. Our mind is able to feel and process our own conscious states. By induction, we are also able to ascribe conscious processing to other human beings. However, once we try to imagine to be another species, as Nagel describes in his seminal work ``What is it like to be a Bat?''\cite{nagel1974like}, we immediately fail to follow such experience consciously.

Another significant issue is that we are not able to determine consciousness by means of behavioural observations as Searle demonstrates in his thought experiment \cite{searle1980minds}. Searle describes a room that we cannot enter. One can pass messages written in Chinese to the room and the room returns messages to the outside world. All the messages and questions passed to the room are answered correctly as a Chinese person would. A first conclusion would be that there is somebody in the ``Chinese Room'' who speaks Chinese and answers the questions. However, the person in the room could also have simply access to a large dictionary that contains all possible questions and the respective answers. When we are not able to understand how the information is actually processed, we will never be able to determine whether a system is conscious or not.

In this article, we want to explore these and different thoughts in literature to address the problem of consciousness. We therefore revisit works in philosophy, neuroscience, artificial intelligence, and machine learning. Following the paradigm of cognitive computational neuroscience \cite{kriegeskorte2018cognitive}, we present how the convergence of these fields could potentially also lead to new insights regarding consciousness.

\section*{The Philosophical Perspective}

More than two thousand years ago, Aristotle was convinced that only humans are endowed with a rational soul. All animals, however, live only with the instincts necessary for survival, like biological automata. Along the same line, in the statement ``Cogito ergo sum'' also Descartes realised being self-aware is reserved for human beings. In his view, this insight is fundamental for any philosophical approach \cite{descartes1990meditations}.

Modern philosophy went on to differentiate the problem into an easy and a hard problem. While the ``easy problem'' is to explain its function, dynamics, and structure, 
the ``hard problem of consciousness'' \cite{chalmers1995facing} is summarised in the Internet Encyclopedia of Philosophy \cite{Weisberg2020} as:

\emph{``The hard problem of consciousness is the problem of explaining why any physical state is conscious rather than nonconscious.  It is the problem of explaining why there is ``something it is like'' for a subject in conscious experience, why conscious mental states ``light up'' and directly appear to the subject.''}

In order to avoid confusion some scientists prefer to speak of ``conscious experience'' or only ``experience'' instead of consciousness \cite{chalmers1995facing}. As already noted, the key problem of deriving models of conscious events is that they can only be perceived subjectively. As such it is difficult to encode such an experience in a way that it can be recreated by others. This gives rise to the so-called ``qualia problem'' \cite{crane2012origins} as we can never be sure, e.g. that the color red consciously looks the same to another person. Extension of this line of thought leads again to Nagel's thought experiment \cite{nagel1974like}.

According to \cite{Weisberg2020}, approaches to tackle the problem from a philosophical point of view are very numerous, but none of them can be considered to be exhaustive:
\begin{itemize}
    \item \textbf{Eliminativism} \cite{rey1988question} demonstrates that the mind is fully functional without the experience of consciousness. Being nonfunctional, consciousness can be neglected. \item The view of 
\textbf{strong reductionism} proposes that consciousness can be deconstructed into simpler parts and be explained by functional processes. Such considerations gave rise to the global work space theory \cite{newman1993neural,baars1994global,baars1994neurobiological} or integrated information theory \cite{tononi2004information,tononi2008consciousness} in neuroscience. The main critique of this view, is that any mechanistic solution to consciousness that is not fully understood will only mimic true consciousness, i.e., one could construct something that appears conscious that simply isn't as the Chinese Room argument demonstrates \cite{searle1980minds}. 
\item \textbf{Mysterianism} proposes that the question of consciousness cannot be tackled with scientific methods. Therefore any investigation is in vain and the explanatory gap cannot be closed \cite{levine2001purple}. 
\item In \textbf{Dualism} the problem is tackled as consciousness being metaphysical that is independent of physical substance \cite{descartes1990meditations}. Modern versions of Dualism exist, but virtually all of them require to reject that our world can be fully described by physical principles. Recently, Penrose and Hammeroff tried to close this gap using quantum theory \cite{penrose1994mechanisms,hameroff2014consciousness}. We dedicate a closer description of this view in a later section of this article. 
\item Assuming that metaphysical world and physical world simply do not interact does not require to reject physics and gives rise to \textbf{Epiphenomenalism} \cite{campbell1992body}. 
\end{itemize}
There are further theories and approaches to address the hard problem of consciousness that we do not want to detail here. To the interested reader, we recommend to study the Internet Encyclopedia of Philosophy \cite{Weisberg2020} as further reading into the topic.

In conclusion, we observe that a major disadvantage of exploring the subject of consciousness by philosophical means is that we are never able to explore the inside of the Chinese Room. Thought alone will not be able to open the black box. Neuroscience, however, offers means to explore the inside by various means of measurement and might offer suitable means to address the problem.

\section*{Consciousness in Neuroscience}

\subsection*{Historical Overview}

In 1924, Hans Berger recorded, for the first time, electrical brain activity using electroencephalography (EEG) \cite{berger1934elektrenkephalogramm}. This breakthrough enabled the investigation of different mental states by means of electrophysiology, e.g. during perception \cite{krauss2018statistical} or during sleep \cite{krauss2018analysis}. The theory of cell assemblies, proposed by Donald Hebb in 1949 \cite{hebb1949organization}, marked the starting point for the scientific investigation of neural networks as the biological basis for perception, cognition, memory, and action. In 1965, Gazzaniga demonstrated that dissecting the corpus callosum which connects the two brain hemispheres with each other results in a split of consciousness \cite{gazzaniga1965observations,gazzaniga2005forty}. Almost ten years later, Weiskrantz et al. discovered a phenomenon for which the term ``blindsight'' has been coined: following lesions in the occipital cortex, humans loose the ability to consciously perceive, but are still able to react to visual stimuli \cite{weiskrantz1974visual,weiskrantz1975blindsight}. In 1983, Libet demonstrated that voluntary acts are preceded by electrophysiological readiness potentials that have their maximum at about $550\,ms$ before the voluntary behavior \cite{libet1983preparation}. He concluded that the role of conscious processing might not be to initiate a specific voluntary act but rather to select and control volitional outcome \cite{libet1985unconscious}. 
In contrast to the above mentioned philosophical tradition from Aristotle to Descartes that consciousness is a phenomenon that is exclusively reserved for humans, in contemporary neuroscience most researchers tend to regard consciousness as a gradual phenomenon, which in principle also occurs in animals \cite{boly2013consciousness}, and several main theories of how consciousness emerges have been proposed so far.

\subsection*{Neural Correlates of Consciousness}

Based on Singer's observation that high-frequency oscillatory responses in the feline visual cortex exhibit inter-columnar and inter-hemispheric synchronization which reflects global stimulus properties \cite{gray1989oscillatory,engel1991interhemispheric,singer1993synchronization} and might therefore be the solution for the so called ``binding problem'' \cite{singer1995visual}, Crick and Koch suggested Gamma frequency oscillations to play a key role in the emergence of consciousness \cite{crick1990towards}. Koch further developed this idea and investigated neural correlates of consciousness in humans \cite{tononi2008neural, koch2016neural}. He argued that activity in the primary visual cortex, for instance, is necessary but not sufficient for conscious perception, since activity in areas of extrastriate visual cortex correlates more closely with visual perception, and damage to these areas can selectively impair the ability to perceive particular features of stimuli \cite{rees2002neural}. Furthermore, he discussed the possibility that the timing or synchronization of neural activity might correlate with awareness, rather than simply the overall level of spiking \cite{rees2002neural}. A finding which is supported by recent neuroimaging studies of visual evoked activity in parietal and prefrontal cortex areas \cite{boly2017neural}. Based on these findings, Koch and Crick provided a framework for consciousness, where they proposed a coherent scheme to explain the neural activation of visual consciousness as competing cellular clusters \cite{crick2003framework}. Finally, the concept of neural correlates of consciousness has been further extended to an index of consciousness based on brain complexity \cite{casarotto2016stratification}, which is independent of sensory processing and behavior \cite{casali2013theoretically}, and might be used to quantify consciousness in comatose patients \cite{seth2008measuring}. 

\subsection*{Consciousness as a Computational Phenomenon}

Motivated by the aforementioned findings concerning the neural correlates of consciousness, Tononi introduced the concept of integrated information, which according to his ``Integrated Information Theory of Consciousness'' plays a key role in the emergence of consciousness \cite{tononi2004information,tononi2008consciousness}. This theory represents one of two major theories of contemporary research in consciousness. According to this theory, the quality or content of consciousness is identical to the form of the conceptual structure specified by the physical substrates of consciousness, and the quantity or level of consciousness corresponds to its irreducibility, which is defined as integrated information \cite{tononi2016integrated}.

Tegmark generalized Tononi's framework even further from neural-net\-work-based consciousness to arbitrary quantum systems. He proposed that consciousness can be understood as a state of matter with distinctive information processing abilities, which he calls ``perceptronium'', and investigates interesting links to error-correcting codes and condensed matter criticality \cite{tegmark2014consciousness, tegmark2015consciousness}.

Even though, there is large consensus that consciousness can be understood as a computational phenomenon \cite{cleeremans2005computational,seth2009explanatory,reggia2016computational,grossberg2017towards}, there is dissent about which is the appropriate level of granularity of description and modeling \cite{kriegeskorte2018cognitive}. Penrose and Hameroff even proposed that certain features of quantum coherence could explain enigmatic aspects of consciousness, and that consciousness emerges from brain activities linked to fundamental ripples in spacetime geometry. In particular, according to their model of orchestrated objective reduction (Orch OR), they hypothesize that the brain is a kind of quantum computer, performing quantum computations in the microtubeles, which are cylindrical protein lattices of the neurons' cytoskeleton \cite{penrose1994mechanisms,hameroff1996orchestrated,hameroff2001biological}. 

However, Tegmark and Koch argue, that the brain can be understood within a purely neurobiological framework, without invoking any quantum-mechanical properties: quantum computations which seek to exploit the parallelism inherent in entanglement, require that the qubits are well isolated from the rest of the system, whereas on the other hand, coupling the system to the external world is necessary for the input, the control, and the output of the computations. Due to the wet and warm nature of the brain, all these operations introduce noise into the computation, which causes decoherence of the quantum states, and thus makes quantum computations impossible. Furthermore, they argue that the molecular machines of the nervous system, such as the pre- and post-synaptic receptors, are so large that they can be treated as classical rather than quantum systems, i.e. that there is nothing fundamentally wrong with the current classical approach to neural network simulations \cite{tegmark2000importance,koch2006quantum,koch2007relation}.

\begin{figure}[tb]
    \centering
    \includegraphics[width=\linewidth]{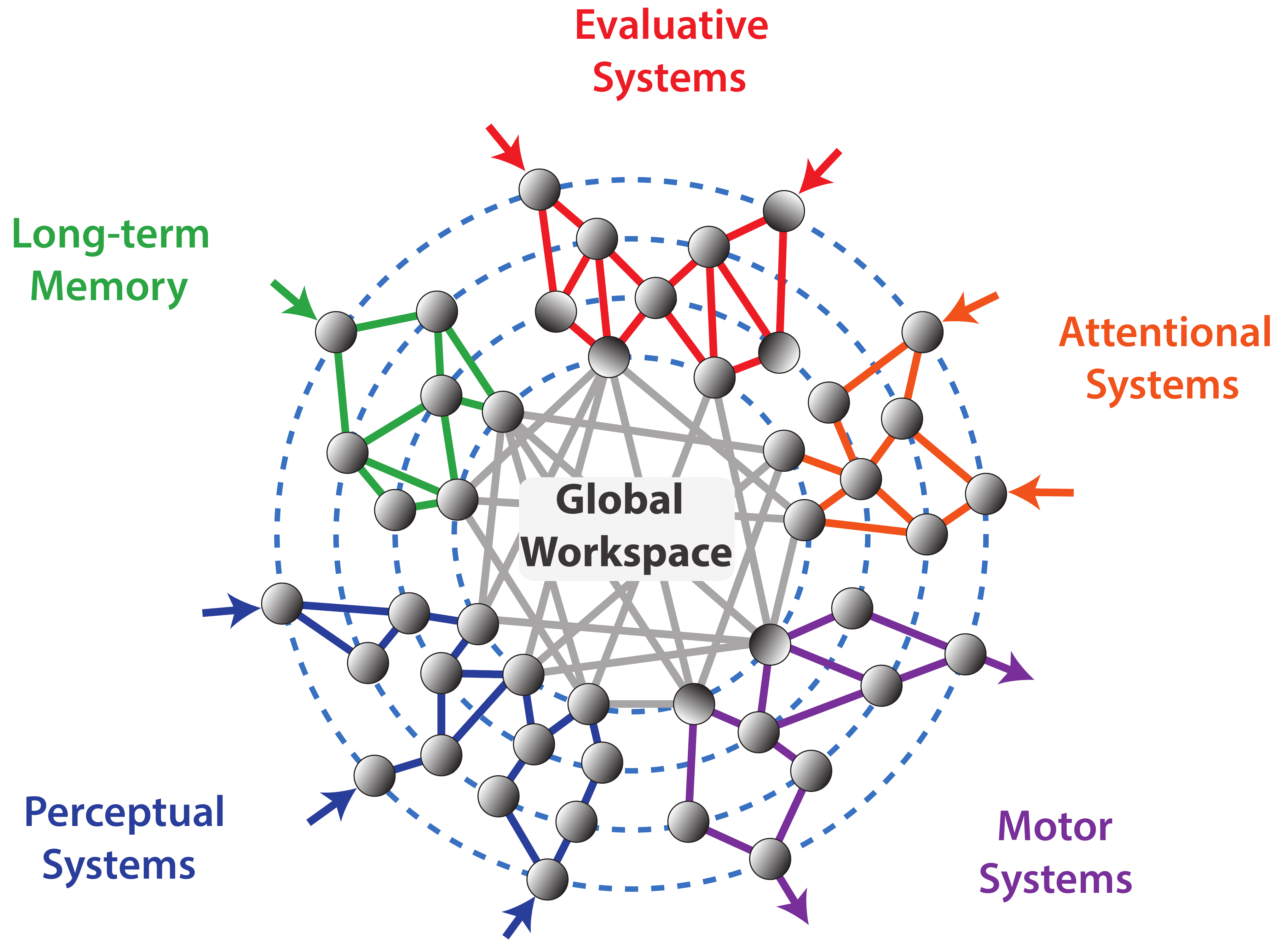}
    \caption{The Global Workspace emerges by connecting different brain areas according to Dehaene.}
    \label{fig:globalworkspace}
\end{figure}

\subsection*{The Global Workspace Theory of Consciousness}

In the 1990s, Baars introduced the concept of a virtual ``Global Workspace'' that emerges by connecting different brain areas (Figure \ref{fig:globalworkspace}) to describe consciousness \cite{newman1993neural,baars1994global,baars1994neurobiological,baars2007global}. This idea was taken up and further developed by Dehaene \cite{dehaene1998neuronal,dehaene2001towards,dehaene2004neural,sergent2004neural,dehaene2011global,dehaene2014toward}. Today, besides the Integrated Information Theory, the Global Workspace Theory represents the second major theory of consciousness, being intensively discussed in the field of cognitive neuroscience. Based on the implications of this theory, i.e., that consciousness arises from specific types of information-processing computations, which are physically realized by the hardware of the brain, Dehaene argues that a machine endowed with these processing abilities \emph{``would behave as though it were conscious; for instance, it would know that it is seeing something, would express confidence in it,would report it to others, could suffer hallucinations when its monitoring mechanisms break down, and may even experience the same perceptual illusions as humans''} \cite{dehaene2017consciousness}.

\begin{figure}[tb]
    \centering
    \includegraphics[width=0.7\linewidth]{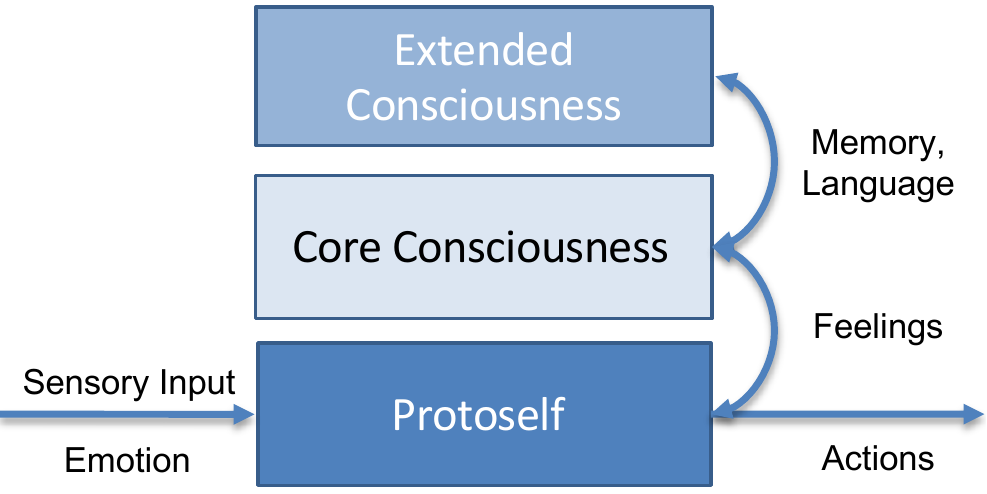}
    \caption{Simplified view of Damasio's model of consciousness: The protoself processes emotions and sensory input unconsciously. Core consciousness arises from the protoself which allows to put the itself into relation. Projections of emotions give rise to higher-order feelings. With access to memory and extended functions such as language processing the extended consciousness emerges.}
    \label{fig:damasiomodel}
\end{figure}

\subsection*{Damasio's Model of Consciousness}

Damasio's model of consciousness was initially published in his popular science book ``The feeling of what happens'' \cite{damasio1999feeling}. Later Damasio also published the central ideas in peer-reviewed scientific literature \cite{damasio2009consciousness}. With the ideas being published first in a popular science book, most publications on consciousness neglect his contributions. However, we believe that his thoughts deserve more attention. Therefore, we want to introduce his ideas quickly in this section.

The main idea in Damasio's model is to relate consciousness to the ability to identify one's self in the world and to be able to put the self in relation with the world. However, a formal definition is more complex and requires the introduction of several concepts first.

He introduces three levels of conscious processing:
\begin{itemize}
    \item The fundamental \textbf{protoself} does not possess the ability to recognize itself. It is a mere processing chain that reacts to inputs and stimuli like an automaton, completely non-conscious. As such any animal has a protoself according to this definition. However, also more advanced lifeforms including humans exhibit this kind of self. 
    \item A second stage of consciousness is the \textbf{core consciousness}. It is able to anticipate reactions in its environment and adapts to them. Furthermore, it is able to recognise itself and its parts in its own image of the world. This enables it to anticipate and to react to the world. However, core consciousness is also volatile and not able to persist for hours to form complex plans.
    
    In contrast to many philosophical approaches, core consciousness does not require to represent representations of the world in words or language. In fact, Damasio believes that progress in understanding conscious processing has been impeded by dependence on words and language. 
    \item The \textbf{extended consciousness} enables human-like interaction with the world. It builds on top of core consciousness and enables further functions such as access to memory in order to create a autobiographic self. Also being able to process words and language falls into the category extended consciousness and can be interpreted as a form of serialisation of conscious images and states. 
\end{itemize}
In Damasio's theory emotion and feelings are fundamental concepts \cite{damasio2001fundamental}. In particular Damasio differentiates emotions from feelings. \textbf{Emotions} are direct signals that indicate a positive or negative state of the (proto-)self. \textbf{Feelings} emerge only in conjunction with images of the world and can be interpreted as a second-order emotion that is derived from the world representation and future possible events in the world. Both are crucial for the emergence of consciousness. Fig.~\ref{fig:damasiomodel} schematically puts the described terms in relation.

After having defined the above concepts, Damasio now goes on to attempt and describe a model of (core) consciousness. In his theory, consciousness does not merely emerge from the ability to identify oneself in the world or an image of the world. For conscious processing, additionally feeling oneself in the sense of desiring to exist is required. Hence, he postulates a feeling, i.e., a derived second-order emotion, between the protoself and its internal representation of the world. Conscious beings as such want to identify oneself in the world and want to exist. From an evolutionary perspective as he argues, this is possibly a mechanism to enforce self-preservation.

In the context of this article, Damasio's theory is interesting for two major reasons. On the one hand, it describes a biologically plausible model of consciousness, as he locates all stages of consciousness to structures in the brain and  associates them to the respective function. On the other hand, Damasio describes a mechanistic model that can at least in theory be completely implemented as a computer program.

Hence, we can conclude that neuroscience is able to describe fundamental processes in the brain that give rise to complex phenomena such as consciousness. However, the means of observation in neuroscience are insufficient. Neither EEG nor fMRI have a temporal and spatial resolution that is even close enough to observe what is happening in the brain in-vivo. At this point, the recent massive progress in artificial intelligence and machine learning comes to our attention and will be the focus of our next section.

\section*{Consciousness in Machine Learning and AI}

In artificial intelligence (AI) numerous theories of consciousness exist \cite{sun2007computational, starzyk2010machine}. Implementations often focus on the global work space theory with only limited learning capabilities \cite{franklin1999software}, i.e. most of the consciousness is hard-coded and not trainable \cite{kotov2017computational}. An exception is the theory by van Hateren which relates the consciousness to close to simultaneous forward and backward processing in the brain \cite{van2019theory}. Yet, algorithms that were investigated so far made use of a global work space and mechanistic hard-coded models of consciousness. Following this line, research on minds and consciousness rather focuses on representation than on actual self-awareness \cite{tenenbaum2011grow}.
Although representation will be important to create human-like minds and general intelligence \cite{gershman2015computational,lake2017building, mao2019neurosymbolic}, a key factor to become conscious is the ability to identify a {\it self} in one's environment \cite{dehaene2017consciousness}. 
A major drawback of pure mechanistic methods, however, is that the complete knowledge on the model of consciousness is required in order to realise and implement them. As such, in order to develop these models to higher forms such as Damasio's extended consciousness, a complete mechanistic model of the entire brain including all connections is required.


\subsection*{Consciousness in Machine Learning}

A possible solution to this problem is machine learning, as it allows to form and train complex models. The topic of consciousness, however, is neglected in the field to a large extent. On the one hand, this is because of the concerns that the brain and consciousness will never be successfully simulated in a computer system \cite{penrose2001consciousness,hameroff2014consciousness}. On the other hand, consciousness is considered to be an extremely hard problem and current results in AI are still meager \cite{brunette2009review}.

The state-of-the-art in machine learning instead focuses on supervised and unsupervised learning techniques \cite{bishop2006pattern}. Another important research direction is reinforcement learning \cite{sutton1998introduction} that aims at learning of suitable actions for an agent in a given environment. As consciousness is often associated with an embodiment, reinforcement learning is likely to be important for modelling of consciousness.

The earliest work that the authors are aware of attempting to model and create agents that learn their own representation of the world entirely using machine learning date back to the early 1990's. Already in 1990, Schmidhuber proposed a model for dynamic reinforcement learning in reactive environments \cite{schmidhuber1990line} and found evidence for self-awareness in 1991 \cite{schmidhuber1991possibility}. The model follows the idea of a global work space. In particular, future rewards and inputs are predicted using a world model.
Yet, Schmidhuber was missing a theory on how to analyse intelligence and consciousness in this approach. Similar to Tononi \cite{tononi2008consciousness}, Schmidhuber followed the idea of compressed neural representation. Interestingly, compression is also key to inductive reasoning, i.e., learning from few examples which we typically deem as intelligent behaviour.

Solomonoff's Universal Theory of Inductive Inference \cite{solomonoff1964formal} gives a theoretic framework to inductive reasoning. It combines information and compression theory and results in a formalisation of Occam's razor preferring simple models over complex ones, as simple models are more likely from an information theoretic point of view  \cite{maguire2016understanding}.  

Under Schmidhuber's supervision, Hutter applied Solomonoff's theory to machine learning to form a theory of Universal Artificial Intelligence \cite{hutter2004universal}. In his theory, intelligent behaviour stems from efficient compression of inputs, e.g. from an environment, such that predictions and actions are performed optimally. Again, models capable of describing a global work space play an important role.

Maguire et al. further expand on this concept to extend Solomonoff's and Hutter's theories to also describe consciousness. Following the ideas of Tononi and Koch \cite{rees2002neural} consciousness is understood as data compression, i.e. the optimal integration of information \cite{maguire2016understanding}. The actual consciousness emerges from binding of information and is inherently complex. As such, consciousness can also not be deconstructed into mechanical sub-components, as the decomposition would destroy the sophisticated data compression. Maguire et al. even provide a mathematical proof to demonstrate that consciousness is either integrated and therefore cannot be decomposed or there is an explicit mechanistic way of modelling and describing consciousness \cite{maguire2016understanding}.

Based on the extreme success of deep learning \cite{lecun2015deep}, also several scientists observed similarities in neuroscience and machine learning. In particular, deep learning allows to build complex models that are hard to analyse and interpret at the benefit of making complex predictions. As such both fields are likely to benefit each other in the ability to understand and interpret complex dynamic systems \cite{marblestone2016toward,van2017computational,hassabis2017neuroscience,kriegeskorte2018cognitive,barrett2019analyzing, richards2019deep, savage2019marriage}. In particular, hard-wiring following biological ideas might help to reduce the search space dramatically \cite{zador2019critique}. This is in line with recent theoretical considerations in machine learning as prior knowledge allows to reduce maximal error bounds \cite{maier2019learning}. Both fields can benefit from these ideas as recent discoveries of e.g. successor representation show \cite{stachenfeld2017hippocampus, gershman2018successor, geerts2019probabilistic}. Several scientists believe that extension of this approach to social, cultural, economic, and political sciences  will create even more synergy resulting in the field of machine behaviour \cite{rahwan2019machine}.

\begin{figure}[tbp]
    \centering
    \includegraphics[width=\linewidth]{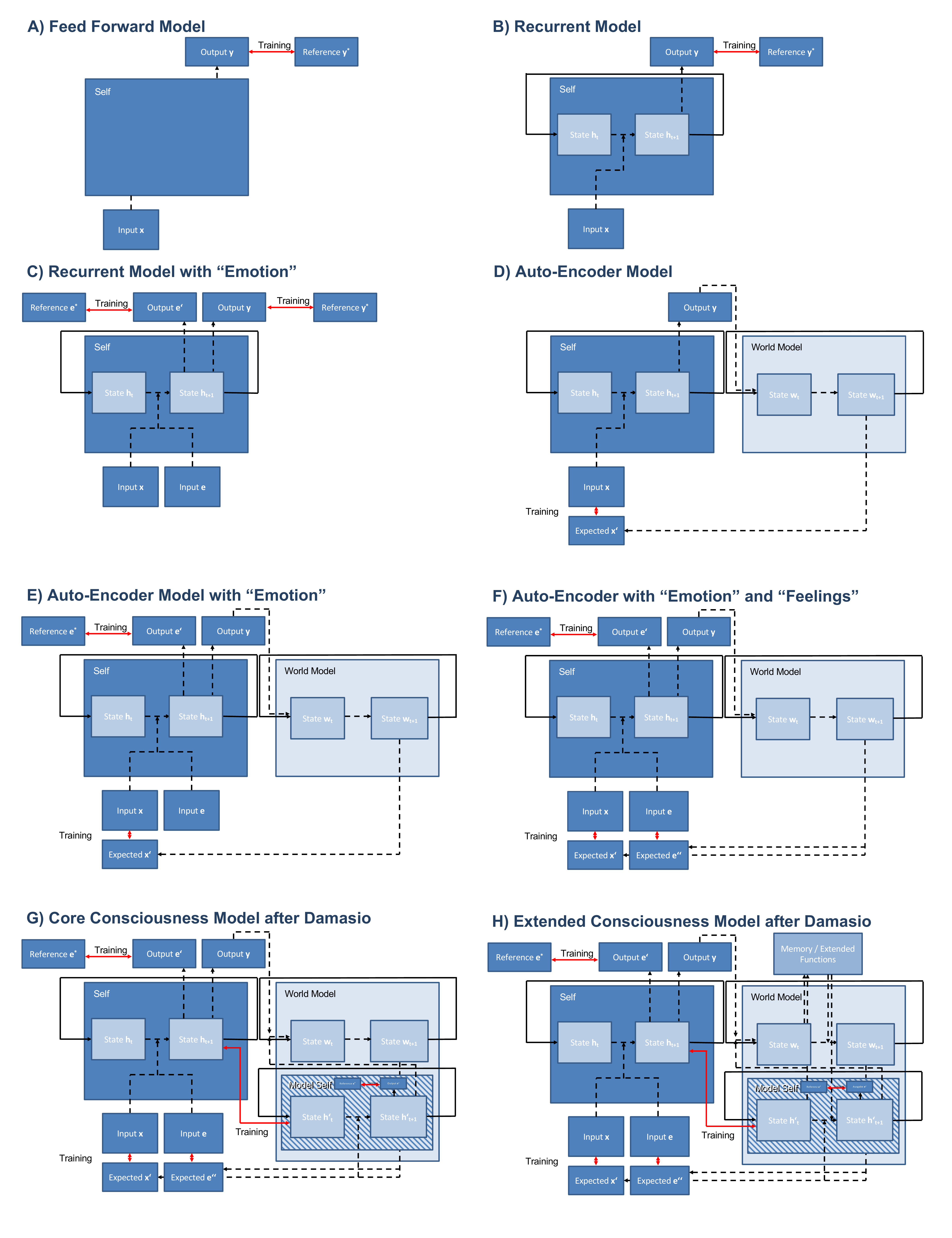}
    \caption{Overview on typical architectures in machine learning. Hard-coded paths, i.e. recurrent connections, are indicated by solid arrows, trainable connections by dashed arrows, and training losses are indicated by red double arrows. While archtiectures A-E do not match theories of consciousness, architectures F-H implement theories by Schmidhuber and Damasio.}
    \label{fig:architectures}
\end{figure}

\section*{Can Consciousness emerge in Machine Learning Systems?}

After having reviewed philosophy, neuroscience, and the state-of-the-art in AI and machine learning, we can now analyse the most important concepts in the field of machine and deep learning to assess whether they have the potential to create consciousness following one of the previous theories. In particular, we focus on the ability of the system to represent a symbol of self and how this self-awareness is constructed, as all theories of consciousness require at least experiencing the self.

We show a brief overview on important architectures in machine learning in Fig.~\ref{fig:architectures}. We denote hard-wired connections as solid arrows to indicate recurrent modes of specific network parts. Dashed lines indicate trainable connections. They could be implemented as a single feed-forward layer, i.e., a universal function approximator. Without loss of generality, they could also be implemented by other deep feed-forward architectures \cite{maier2019gentle} and could thus be inherently complex. Red double arrows indicate training losses to adjust the trainable weights of the dashed arrows. 

Fig.~\ref{fig:architectures} A shows a simple feed-forward architecture that requires external labelled training data $\bf{y}^*$ to adjust its trainable weights given input $\bf{x}$ to produce output $\bf y$. Fig.~\ref{fig:architectures} B shows a similar setup, for the recurrent case. Note that we only indicate a simple recurrent cell here with time-dependent state $\bf h_t$. Without loss of generality, this could also be realised using gated recurrent units \cite{cho2014properties} or long short term memory cells \cite{hochreiter1997long}. Both models fall into the category of supervised learning. Consciousness is not supported by any of the theories presented so far.

In Fig.~\ref{fig:architectures} C, we introduce the concept of ``Emotion'' following Damasio's wording. In machine learning terms, this reflects an additional loss. Now, the system receives an additional input $\bf e$ that is associated to a valence or value. Without loss of generality, we can assume positive entries in $\bf e$ to be associated to desirable states for the system and negative values to undesirable states. As such, training using $\bf e$ falls into the category of reinforcement learning that aims at maximizing future values of $\bf e$. In order to model competing interests and saturation effects, e.g. a full battery does not need to be charged further, we introduce a reference $\bf e^*$ that is able to model such effects. Note that we deem the system to be able to predict the expected future reward $\bf e'$ from its current state $\bf h_t$ following a deep Q learning paradigm \cite{mnih2015human}. Here we use $\bf e'$ and $\bf e^*$ to construct a trainable reinforcement loss, to be able to learn from low-level rewards/emotions $\bf e$. Although being able to learn, the system still needs supervision to train the weights producing output $\bf y$ using a reference. This setup also does not match any theory of consciousness so far.

As self-awareness is a requirement for base consciousness, we deem a world model to be necessary. Such an approach his shown in Fig.~\ref{fig:architectures} D. Given from the produced output $\bf y$, the world model is used to create an expected future input $\bf x'$. In the figure, we chose a recurrent model capturing the state of the world in $\bf w_t$ that is independent of the internal state of the actual agent $\bf h_t$. To gain consciousness, this model misses at least a link from internal to external state and ``emotions'' that would guide future decisions. Adding low-level rewards/emotions results in Fig.~\ref{fig:architectures} E. Again world and self are disconnected inhibiting self-representation and self-discovery. Approaches like this are already being explored for video game control \cite{kaiser2019model}.

With a world-model being present, we are now able to predict future rewards $\bf e''$ that also take into account the state of the world and the chosen action. As such Fig.~\ref{fig:architectures} F is the first one that would implement a trainable version of deep Q learning. Development of consciousness is debatable, as the model does not feature a link between the state of the world and the state of the agent. If we would add a trainable connection from $\bf h_t$ to $\bf w_t$ and vice versa, we would end up with Schmidhuber's Model from 1990 \cite{schmidhuber1990line} for which Schmidhuber found evidence to develop self representation \cite{schmidhuber1991possibility}.

Interestingly, Damasio's descriptions follow a similar line in \cite{damasio1999feeling}. We depict a model implementing Damasio's core consciousness in Fig.~\ref{fig:architectures} G. 
As Schmidhuber, Damasio requires a connection from the world model $\bf w_t$ to the body control system $\bf h_t$. However, in his view, consciousness does not emerge by itself. It is enforced by a ``feeling'' that is expressed as a loss in the world of machine learning. As such, the Damasio model of core consciousness requires a loss that aims at the recovery of the image of the self in the world model. If this is implemented as a loss, we are able to express the desire to exist in the world. If implemented merely as trainable weights, we arrive at the theory of integrated information that creates consciousness as maximized compression of the world, the self, and their interaction. Interestingly, these considerations also allow integration of attention \cite{vaswani2017attention} and other concepts of resolving context information in machine learning. Realised in a biological learning framework, e.g. using neuromodulators like Dopamin \cite{russek2017predictive}, the notion of loss and connection will disappear and the models of Damasio, Schmidhuber, Tononi, Koch, and Dehane turn out to be different descriptions of the same principles.

Note that the models of consciousness that we have discussed so far are very basic. They do not concern language, memory, or other complex multimodal forms of processing, planning, induction, or representation. Again, we follow Damasio at this point in Fig.~\ref{fig:architectures} H in which all of these sophisticated processes are mapped into a block ``Memory / Extended Functions''. Note although we omit these extended functions, we are able to integrate them using trainable paths. As such, the model of core consciousness acts as a ``neural operating system'' that is able to update 
also higher order functions according to the needs of the environment. With increasing ``extended functions'', the degree of complexity and ``integrated information'' rises measurably as also observed by Casarotto \cite{casarotto2016stratification}.

This brings us back to the original heading of our section: There are clearly theories that enable modelling and implementation of consciousness in the machine. On the one hand, they are mechanistic to the extend that they can be implemented in programming languages and require similar inputs as humans would do. On the other hand, even the simple models in Fig.\ref{fig:architectures} are already arbitrarily complex, as every dashed path in the models could be realised by a deep network. As such also training will be hard. Interestingly, the models follow a bottom-up strategy such that training and development can be performed in analogy to biological development and evolution. The models can be trained and grown to more complex tasks gradually.

\section*{Discussion}

Existence of consciousness in the machine is a hot topic of debate. Even with respect to the simple core consciousness, we observe opinions ranging from ``generally impossible'' \cite{carter2018conscious} through ``plausible'' \cite{dehaene2017consciousness} to ``has already been done'' \cite{schmidhuber1991possibility}. Obviously, all of the suggested models cannot solve the qualia problem or the general problem on how to demonstrate whether a system is truly conscious. All of the emerging systems could merely be mimicking conscious behaviour without being conscious at all (even Fig.~\ref{fig:architectures} A). Yet as already discussed by Schmidhuber \cite{schmidhuber1991possibility}, we would be able to measure correlates of self recognition similar to neural correlates of consciousness in humans \cite{koch2016neural} which could help to understand consciousness in human beings. However, as long as we have not solved how to provide proof of consciousness in human beings, we will also fail to do so in machines as the experience of consciousness is merely subjective. 

Koch and Dehaene discussed the theories of global work space and integrated information as being opposed to each other \cite{carter2018conscious}. In the models found in Fig.~\ref{fig:architectures}, we see that both concepts require a strong degree of interconnection. As such, we do not see why both concepts are fundamentally opposing. A global work space does not necessarily have to be encoded in decompressed state. Also, Maguire's view of integrated information \cite{maguire2016understanding} is not necessarily impossible to implement mechanistically, as we are able to use concepts of deep learning to train highly integrated processing networks. In fact, as observed by neuroscience \cite{kriegeskorte2018cognitive}, both approaches might support each other yielding methods to construct and reproduce biological processes in a modular way. This allows the integration of representation \cite{gershman2015computational} and processing theories \cite{sun2007computational} as long as they can be represented in terms of deep learning compatible operations \cite{maier2019learning}.  




In all theories that we touched in this article, the notion of self is fundamental. Hence, all presented theories of consciousness require embodiment as basis for consciousness. As such, a body is required for conscious processing. Also the role of emotion and feelings is vital. Without emotion or feelings, the system cannot be trained and thus it can not adopt to new environments and changes of circumstances. In the machine learning inspired models, we assume that a disconnection between environment and self would cause a degradation of the system similar to the one that is observed in human beings in locked-in state \cite{kubler2008brain}. This homeostatsis was also deemed important by Man et al. \cite{man2019homeostasis}.

Similar to the problems identified by Nagel, also the proposed mechanistic machine learning models will not be able to understand ``what it is like'' to be a bat. However, the notion of train-/learnable programs and connections or adapters might offer a solution to explore this in the future. Analogously, one cannot describe to somebody ``what it is like'' to play the piano or to snowboard on expert level unless one really acquires the ability. As such also the qualia problem persists in machine consciousness. However, we are able to investigate the actual configuration of the representation in the artificial neural net offering entirely new mechanisms of insight.

In Damasio's theory, consciousness is effectively created by a training loss that causes the system to ``want'' to be conscious, i.e., "Cogito ergo sum" becomes "Sentio ergo sum". Comparison between trainable connections after \cite{schmidhuber1990line}, attention mechanisms \cite{vaswani2017attention}, and this approach are within the reach of future machine learning models which will create new evidence for the discussion of integrated information and global work spaces. In fact, Schmidhuber has already taken up the work on combination of his early ideas with modern approaches from deep learning \cite{schmidhuber2015learning,schmidhuber2018one}. 

With models for extended consciousness, even the notion of the Homunculus \cite{kenny2016homunculus} can be represented by extension of the self with another self pointer. In contrast to common rejection of the Homunculus thought experiment, this recurrent approach can be trained using end-to-end systems comparable to AlphaGo \cite{silver2016mastering}.

Damasio also presents more interesting and important work that is mostly omitted in this article for brevity. In \cite{damasio1999feeling} he also relates structural brain damage to functional loss of cognitive and conscious processing. Also the notion of emotion is crucial in a biological sense and is the driving effect of homeostasis. In \cite{man2019homeostasis}, Damasio already pointed out that this concept will be fundamental for self-regulating robotic approaches.

With the ideas of cognitive computational neuroscience \cite{kriegeskorte2018cognitive} and the approaches detailed above, we will design artificial systems that approach the mechanisms of biological systems in an iterative manner. With the iterations, the artificial systems will increase in complexity and similarity to the biological systems. However, even if we arrive at an artificial system that performs identical computations and reveals identical behaviour as the biological system, we will not be able to deem this system as conscious beyond any doubts. The true challenge in being perceived as conscious will be the acceptance by human beings and society. As Alan Turing already proposed in his imitation game in 1950 to decide whether a machine is intelligent or even conscious \cite{turing1950computing}, the ascription of such by other humans is a critical factor. For this purpose, Turing's Test has already been extended to also account for embodiment \cite{french2012moving}. However, such tests are only necessary, but not sufficient as Gary Marcus pointed out: Rather simple chat bot models are already able to beat Turing's Test in some occasions \cite{vardi2014would}. Hence, requirements for conscious machines will comprise, the similarity to biological conscious processes, the ability to convince human beings, and even the machine itself. As such, we deem it necessary to look as some ethical implications at this point.


\section*{Ethical implications}

Being able to create systems that are indistinguishable from conscious beings that are potentially conscious also raises ethical concerns. First and foremost, in the transformation from core consciousness to extended consciousness, the systems gain the ability to link new program routines. As such systems following such a line of implementation need to be handled with care and should be experimented on in a contained environment. With the right choice of embodiment in a virtual machine or in a robotic body, one should be able to solve such problems.

Of course there are also other ethical concerns, the more we approach human-like behaviour. A first set of robotic laws has been introduced in Asimov's novels \cite{clarke1993asimov}. Even Asimov considered the rules problematic as can be seen from the plot twists in his novels. Aside this, being able to follow the robotic laws requires the robot to understand the concepts of ``humans'', ``harm'', and ``self''. Hence, such beings must be conscious. Therefore, tampering with their memories, emotions, and feelings is also problematic by itself. Being able to copy and reproduce the same body and mind does not lead to further simplification of the issue and implies the problem that we have to agree on ethics and standards of AI soon \cite{jobin2019global}.

\section*{Conclusion}

In this article, we reviewed the state-of-the-art theories on consciousness in philosophy, neuroscience, AI, and machine learning. We find that all three disciplines need to interact to push research in this direction further. Interestingly, basic theories of consciousness can be implemented in computer programs. In particular, deep learning approaches are interesting as they offer the ability to train deep approximators that are not yet well understood to construct mechanistic systems of complex neural and cognitive processes. We reviewed several machine learning architectures and related them to theories of strong reductionism and found that there are neural network architectures from which base consciousness could emerge. Yet, there is still a long way to form  human-like extended consciousness.


\section*{Data availability}
All data in this publication are publicly available. 

\section*{Code availability}
All software used in this article is publicly available.

\section*{Acknowledgments}
This work was funded by the Deutsche Forschungsgemeinschaft (DFG, German Research Foundation): grant KR5148/2-1 to PK -- project number 436456810, the Emergent Talents Initiative (ETI) of the University Erlangen-Nuremberg (grant 2019/2-Phil-01 to PK). Furthermore, the research leading to these results has received funding from the European Research Council (ERC) under the European Union's Horizon 2020 research and innovation programme (ERC grant no. 810316).

\section*{Author contributions}
PK and AM contributed equally to this work.

\section*{Ethics declarations}
\subsection*{Competing interests}
The authors declare no competing financial interests.


\FloatBarrier
\bibliographystyle{unsrt}
\bibliography{literature}


\end{document}